\documentclass[10pt,twocolumn,letterpaper]{article}

\usepackage[pagenumbers]{cvpr} % To force page numbers, e.g. for an arXiv version

\definecolor{cvprblue}{rgb}{0.21,0.49,0.74}
\usepackage[pagebackref,breaklinks,colorlinks,allcolors=cvprblue]{hyperref}
\usepackage{stfloats}
\usepackage{capt-of}
\usepackage{graphicx}
\usepackage{makecell}
\usepackage{pifont}
\usepackage{pifont}
\usepackage{multirow}
\usepackage{tabularx}
\usepackage{cuted}
\usepackage{capt-of}
\usepackage{siunitx}
\usepackage[most]{tcolorbox}

\def\paperID{} % *** Enter the Paper ID here
\def\confName{CVPR}
\def\confYear{2026}

\title{InstAP:~\underline{Inst}ance-\underline{A}ware Vision-Language~\underline{P}re-Train for Spatial-Temporal Understanding}

\author{Ashutosh Kumar,~~Rajat Saini,~~Jingjing Pan,~~Mustafa Erdogan,~~Mingfang Zhang,\\~~Betty Le Dem,~~Norimasa Kobori,~~Quan Kong\\
Woven by Toyota\\
{\tt\small \{firstname.lastname\}@woven.toyota}
}

\begin{document}

\twocolumn[{
\begin{center}
  \maketitle
  \includegraphics[width=0.92\textwidth]{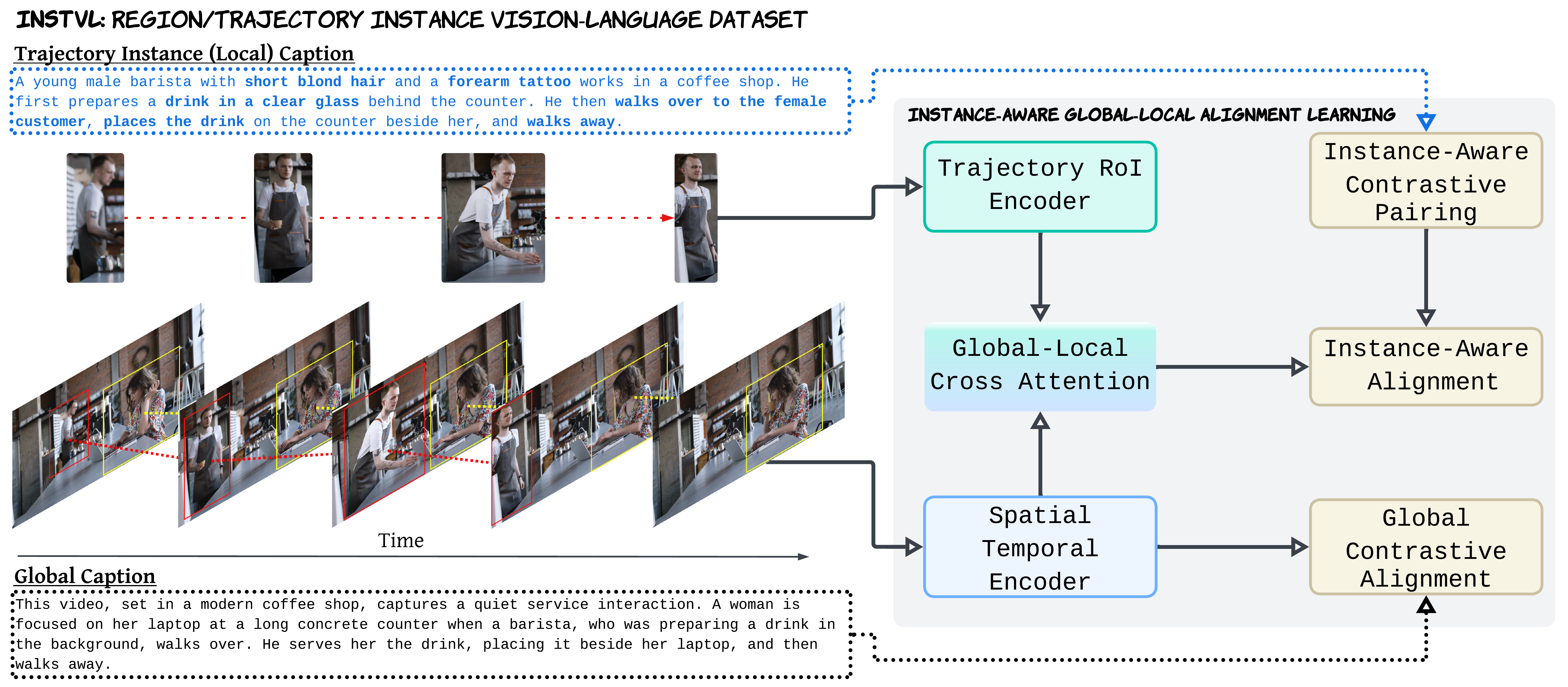}
  % \captionof{figure}{The motivation and key contributions of our proposal: 1) A new dataset \textit{InstVL} is proposed with region and instance trajectory level detailed captions for the objects in the given videos, along with the global detailed captions for describing the entire video/image; 2) Based on \textit{InstVL}, we propose an Instance-aware Global-Local Spatial-Temporal Alignment Learning as a pre-train approach that could learn the local fine-grained visual representation and the global video representation simultaneously with global-local attention under contrastive alignment for fine-grained spatial-temporal understanding.}
  % \captionof{figure}{
  %  Conceptual overview of our \textbf{InstAP} framework and \textbf{InstVL} dataset.
  % \textbf{Left:} The InstVL dataset provides dual-granularity annotations for a video: a holistic \textit{Global Caption} and a fine-grained \textit{Trajectory Instance Caption} grounded to a specific entity (e.g., the barista).
  % \textbf{Right:} The InstAP processes this data with a \textit{Spatial Temporal Encoder} for global context and a \textit{Trajectory ROI Encoder} for instance details. These are fused via \textit{Global-Local Cross Attention} and jointly optimized with both \textit{Global Alignment} and \textit{Instance-Aware Alignment} objectives.
  % }

  
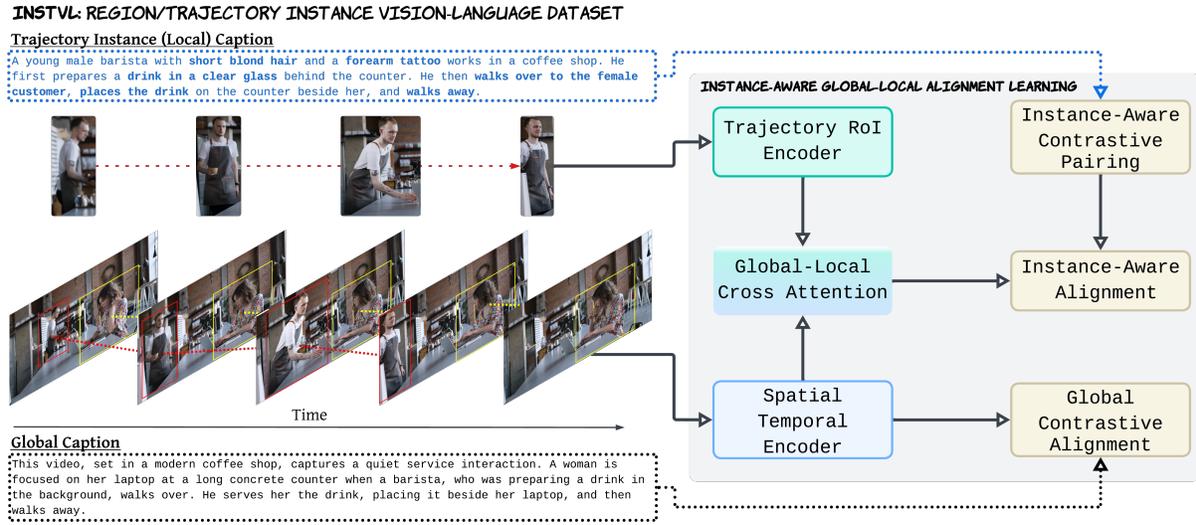
\captionof{figure}{Conceptual overview of the \textbf{InstAP} framework and \textbf{InstVL} dataset. \textbf{Left:} InstVL features dual-granularity video annotations: holistic \textit{Global Captions} and entity-grounded \textit{Trajectory Instance Captions}. \textbf{Right:} InstAP fuses global and instance-level features via \textit{Global-Local Cross Attention}, optimizing through joint \textit{Global} and \textit{Instance-Aware Alignment} objectives.}
  
  \label{fig:concept}
\end{center}
}]
\begin{abstract} Current vision-language pre-training (VLP) paradigms excel at global scene understanding but struggle with instance-level reasoning due to global-only supervision. We introduce \textbf{InstAP}, an \textbf{Inst}ance-\textbf{A}ware \textbf{P}re-training framework that jointly optimizes global vision-text alignment and fine-grained, instance-level contrastive alignment by grounding textual mentions to specific spatial-temporal regions. To support this, we present \textbf{InstVL}, a large-scale dataset ($2$ million images, $50,000$ videos) with dual-granularity annotations: holistic scene captions and dense, grounded instance descriptions. On the InstVL benchmark, InstAP substantially outperforms existing VLP models on instance-level retrieval, and also surpasses a strong VLP baseline trained on the exact same data corpus, isolating the benefit of our instance-aware objective. Moreover, instance-centric pre-training improves global understanding: InstAP achieves competitive zero-shot performance on multiple video benchmarks, including MSR-VTT and DiDeMo. Qualitative visualizations further show that InstAP localizes textual mentions to the correct instances, while global-only models exhibit more diffuse, scene-level attention. \end{abstract}  
\section{Introduction}
\label{sec:intro}

Vision-Language Pre-training has fundamentally reshaped the landscape of representation learning, moving beyond supervised learning on fixed category datasets. Seminal work in the image domain, notably CLIP \cite{radford2021learning}, demonstrated the power of learning transferable visual representations directly from natural language supervision. By employing contrastive learning objectives on hundreds of millions of image-text pairs harvested from the web, CLIP learned representations capable of impressive zero-shot generalization across diverse visual concepts, significantly broadening the scope compared to traditional classification-based pre-training. This success spurred intense interest in extending VLP to the video domain, a naturally richer but substantially more complex modality.

Most existing approaches focus on capturing global, coarse-grained correspondences between an entire video and its caption~\cite{xu2021videoclip,bansal2024videocon, tian2024holistic, yang2024dgl, li2023prototype, sun2022long, lin2022egocentric, zellers2021merlot}, often neglecting fine-grained instance-level semantics. This leaves a critical gap: models struggle to identify and distinguish specific objects or entities mentioned in text. For example, given a caption \emph{“a child throws a red ball while a dog jumps”}, a model trained only on global alignments might grasp the overall event but fail to localize which visual region corresponds to the \emph{“ball”} or the \emph{“dog”}. Such shortcomings in instance-level understanding can limit performance on downstream tasks that require precise grounding of language in video, including fine-grained retrieval, spatial-temporal grounding, and object-centric question answering.

Learning fine-grained, instance-aware representations is non-trivial. On one hand, most large-scale video-text datasets provide only high-level descriptions, lacking the grounded annotations necessary to learn instance-word correspondences. On the other hand, prevailing pre-training objectives reward holistic video-text alignment, providing little incentive for the model to attend to subtle, instance-specific details. While recent works have attempted to address this by grafting instance-level cues onto models post-hoc, they often rely on pre-trained object detectors \cite{li2020oscar,zhang2021vinvl} or auxiliary specialization heads \cite{wang2025internvideo2,yuan2024osprey}. These signals are often treated as auxiliary features rather than being integrated into the core representation learning, inheriting detector errors and failing to achieve true instance-level alignment. Consequently, a general and effective solution for instance-aware video pre-training remains elusive.

In this paper, we propose \textbf{InstAP}, an \textbf{Inst}ance-\textbf{A}ware vision-language \textbf{P}re-training framework (Fig.~\ref{fig:concept}) that learns representations capturing both global context and rich instance-level information. Instead of aligning only whole video clips with captions, InstAP introduces an instance-centric training objective that enforces alignment between specific textual mentions and their corresponding object-level visual features. This guides the model to ground individual entities, making the learned representations highly discriminative at the instance level while preserving holistic semantic understanding. To enable this training, we introduce \textbf{InstVL}, a large-scale dataset of $2$ million images and $50,000$ videos with dual-granularity annotations: a holistic scene caption and dense, grounded instance-level descriptions.

Our experiments demonstrate three key contributions and findings: 
\begin{itemize} 
    \item We introduce the InstVL dataset and the InstAP framework, which significantly outperforms existing models. By surpassing a strong VLP baseline trained on the same corpus, we demonstrate that InstAP's gains stem from our instance-aware alignment framework rather than just data scaling.
    \item InstAP achieves competitive generalization on zero-shot benchmarks like MSR-VTT and DiDeMo, proving that fine-grained alignment across instance and global levels actually enhances holistic scene understanding.
    \item Qualitative analysis confirms InstAP's ability to precisely ground textual phrases to visual instances, a capability notably absent in traditional global-only models.
\end{itemize}
\section{Related Works}
\label{sec:rw}

\subsection{Grounded Vision-Language Datasets}
\label{sec:rw_datasets}

A core bottleneck for instance-aware pre-training has been the lack of appropriate, large-scale training data. While image-domain datasets like Visual Genome~\cite{krishna2017visual} and Flickr30k Entities~\cite{plummer2015flickr30k} provide region-level annotations, they are limited to grounding structured attributes or short phrases, not the full, free-form sentences needed for generative understanding. This gap is more severe in the video domain. Datasets with rich spatial-temporal trajectories are often highly domain-specific; for example, in autonomous driving~\cite{caesar2020nuscenes}, efforts to add captions have relied on rule-based, \textit{template-generated} language~\cite{gopinathan2025temporal}, which lacks linguistic diversity. Conversely, general-purpose video datasets that provide trajectories, like VidOR~\cite{shang2019annotating}, are limited to closed-vocabulary, structured predicates (\emph{e.g.}, \texttt{<subject,chase,object>}). Finally, other general datasets like ActivityNet-Entities~\cite{zhou2019grounded} only ground \textit{noun phrases} to a \textit{single, static frame}, failing to capture temporal continuity. The \textbf{InstVL} corpus is developed to fill this critical gap, providing the first large-scale, general-domain resource with free-form sentence annotations for both static regions and full video trajectories.

\subsection{Image-Language Pre-training}

The foundation for modern vision-language understanding was largely established in the image domain. Seminal work, notably CLIP \cite{radford2021learning}, showcased how contrastive pre-training on web-scale image-text data could yield transferable visual representations with remarkable zero-shot performance. Subsequent work refined this paradigm, \emph{e.g.}, with alternative loss functions~\cite{zhai2023sigmoid,tschannen2025siglip}. Concurrently, other works pushed for richer localization by incorporating region-level objectives~\cite{li2022grounded, zeng2021multi, zhong2022regionclip}. This evolution demonstrates a move from global-only alignment towards capturing finer-grained semantics. Our work builds on this insight, extending the pursuit of fine-grained understanding to the spatial-temporal dynamics of video.

\subsection{Video-Language Pre-training}

Extending VLP to video required addressing temporal modeling and computational complexity. Many models~\cite{yan2022multiview, xu2021videoclip} adapted the CLIP paradigm, aligning entire video clip embeddings with text. While successful for global retrieval, these methods inherently average features, suppressing instance-level details. A second branch leverages self-supervised objectives, such as reconstructing masked portions~\cite{tong2022videomae}, inspired by BERT~\cite{devlin2019bert} and MAE~\cite{he2022masked}. Recent work like UMT~\cite{li2023unmasked} and VideoPrism~\cite{zhao2024videoprism} advanced this by distilling from a CLIP teacher to a video student. While innovations like semantic masking might implicitly focus on salient objects, the alignment target remains the teacher's global representation, an indirect signal that itself lacks instance-specific grounding. Ultimately, both frameworks learn representations where instance-level cues are, at best, emergent and implicit, not explicitly modeled or aligned with specific textual mentions.

\subsection{Towards Instance-Level Understanding in Vision-Language}

Limitations of global-only models motivated efforts to inject finer-grained information. A dominant strategy is adding locality post-hoc via detector-based methods~\cite{li2020oscar,zhang2021vinvl,li2022grounded} that feed in region tags, coupling performance to detector quality. A recent variant adds specialized modules, \emph{e.g.}, instance-segmentation heads~\cite{wang2025internvideo2,yuan2024osprey}. While successful, these treat instance understanding as an auxiliary specialization, not a core encoder capability. Detector-free, region-phrase mining methods~\cite{zeng2021multi,zhong2022regionclip,li2022fine} have shown promise on images but have not scaled effectively to video pre-training.

A critical gap remains: embedding instance awareness directly into large-scale video pre-training. Our work fundamentally departs from these ``grafted-on'' solutions. We posit that instance-level comprehension must be a core property of the representation, not an auxiliary task. We therefore introduce InstAP, a framework that embeds instance-awareness directly into the pre-training phase, learning a unified representation for both holistic and instance-level understanding.
\section{Methodology}
\label{methodology}

\subsection{InstVL dataset}
\label{instvl_dataset}

\begin{figure}
    \centering
    \includegraphics[width=\linewidth]{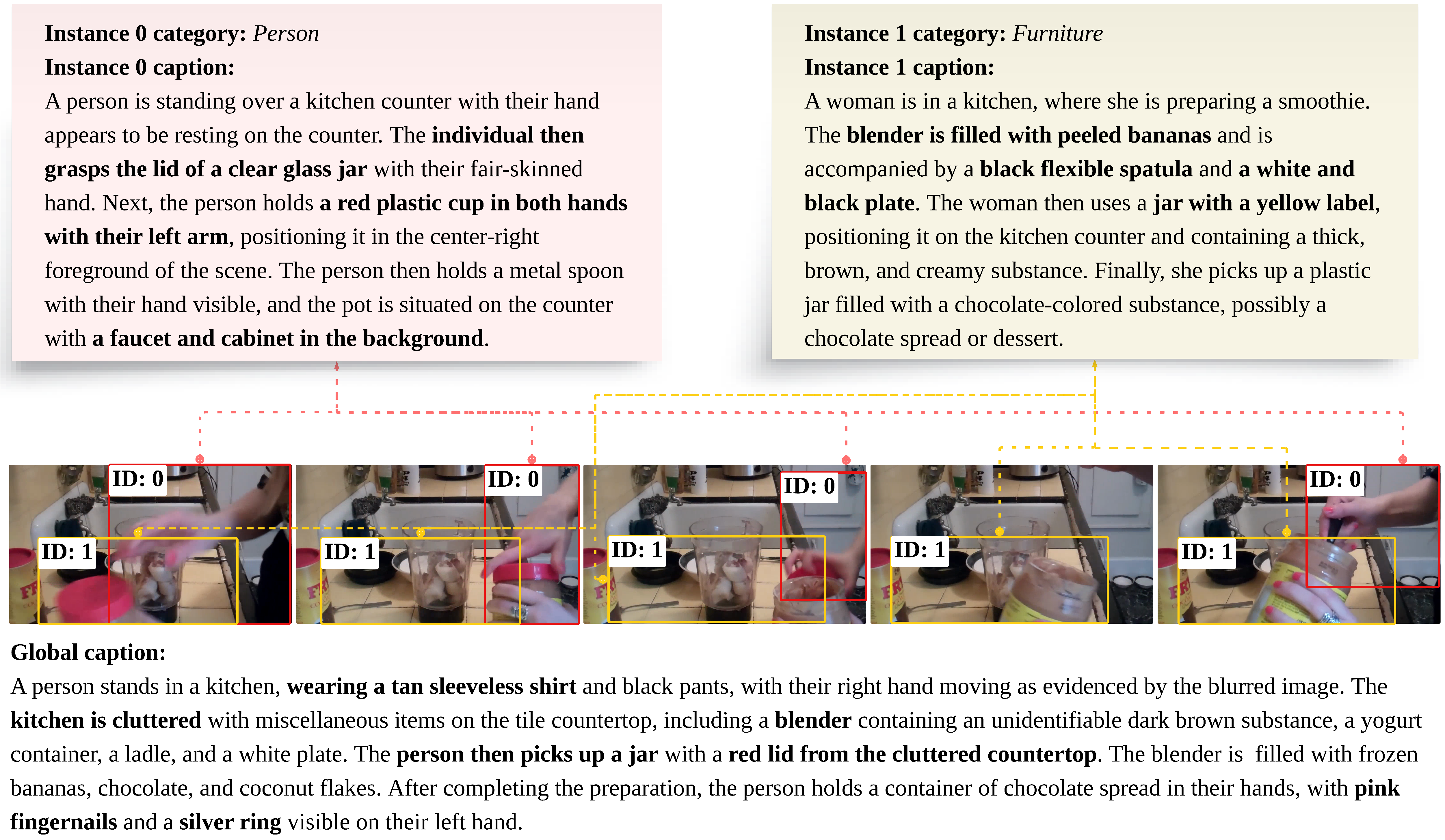}
      \caption{Illustration of our \textbf{InstVL} dataset. We display sampled frames with color-coded, temporally-consistent instance trajectories (\textit{e.g.}, ID: 0, ID: 1). The top text provides the fine-grained \textit{instance captions} grounded to these trajectories; the bottom text provides the holistic \textit{global caption} for the entire scene.}
\label{fig:ia_pt_dataset}
\end{figure}

The \textbf{InstVL} corpus is a new large-scale vision-language dataset, containing $2$ million images and $50,000$ video clips, designed to facilitate instance-aware pre-training. Its key contribution is the dual-granularity textual annotations provided for each visual sample: (1) a scene caption for holistic context and (2) a collection of instance-level captions grounded in specific visual regions (for images) or spatial-temporal trajectories (for videos), as illustrated in Fig.~\ref{fig:ia_pt_dataset}.
\subsubsection{Data Curation Pipeline}

Our main training dataset images are drawn from LAION-400M~\cite{schuhmann2021laion}, while the videos are sourced from processed segments of HDVILA~\cite{xue2022hdvila}. To create our zero-shot test splits, we exclusively used images from COYO~\cite{kakaobrain2022coyo-700m}, ensuring no overlap with the training sources. We first processed videos with AutoShot~\cite{zhu2023autoshot} for scene segmentation. Next, we generated spatial-temporal instance groundings using GroundingDINO~\cite{liu2024grounding} as an open-vocabulary detector and SAM2~\cite{ravi2024sam2segmentimages} for instance tracking. To generate the dual-granularity text, we fed these regions and trajectories with visual bounding box prompts to a large vision-language model~\cite{hurst2024gpt}, which generated both the holistic scene captions and the fine-grained instance-level descriptions. This pipeline underwent several iterations of manual human checking to refine the prompting techniques and ensure high-quality, descriptive annotations. Each image or video sample contains: (1) a single scene caption and (2) a variable number of instance annotations. For images, an instance is a $2$D bounding box. For videos, it is a temporal trajectory of boxes. Each instance annotation is coupled with a free-form sentence describing its specific appearance, attributes, or actions.

\subsubsection{InstVL Test Suite}
To facilitate systematic benchmarking, we curate a held-out test suite
with five mutually exclusive subsets: \texttt{InstVL-1K (img)} and
\texttt{InstVL-10K (img)} for images, \texttt{InstVL-1K (img-zero)} and
\texttt{InstVL-10K (img-zero)} for zero-shot images, and
\texttt{InstVL-1K (video)} for videos.
The \texttt{InstVL-1K (img-zero)}\slash\texttt{InstVL-10K (img-zero)} subsets
are sourced entirely from COYO, whereas the main training images
(and their corresponding test splits) are from LAION. This introduces a distribution shift that lets us confirm that our model’s
performance is not merely inherited from the training distribution.

\begin{figure*}
    \centering
    \includegraphics[width=0.9\linewidth]{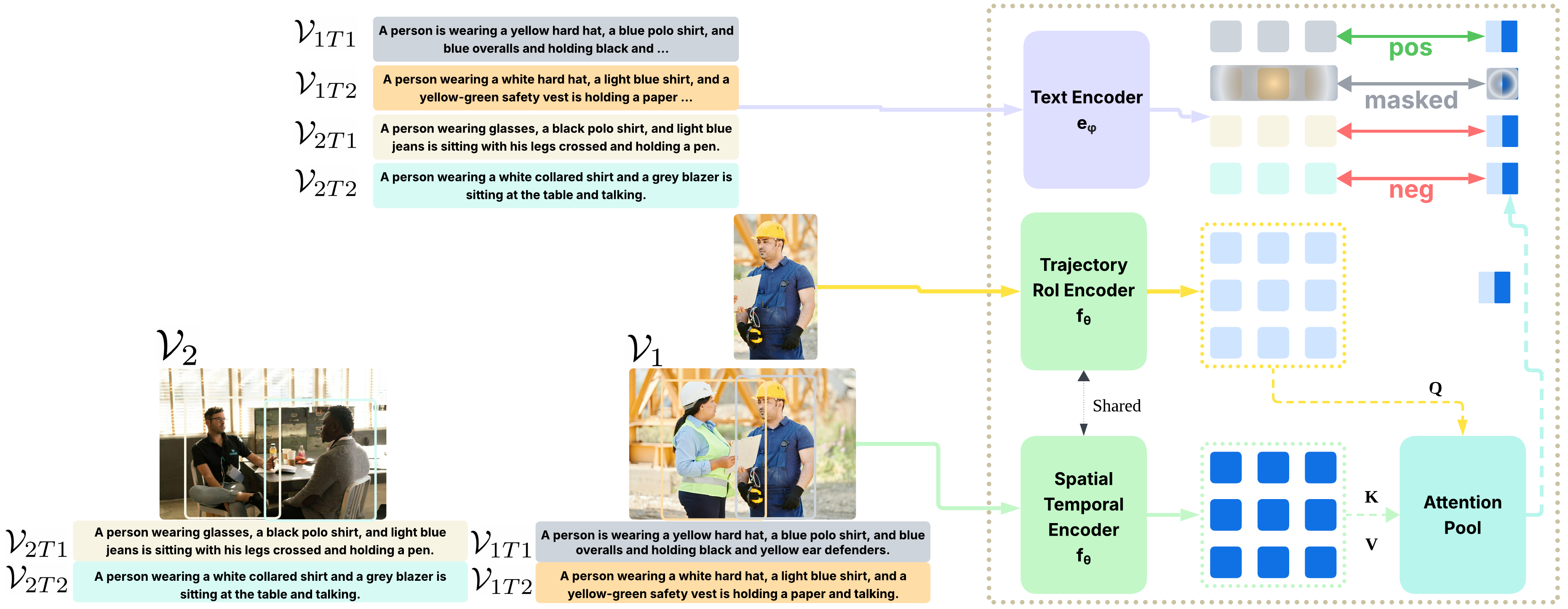}
   \caption{Our instance-aware alignment mechanism. Instance features (Query $Q$) from a Trajectory RoI Encoder ($f_{\theta}$) are fused with global context (Key $K$, Value $V$) via an Attention Pool to create an instance-aware embedding. This embedding is contrasted with text features ($e_{\phi}$). The loss forces the model to match positive pairs ($V_{1T1}$) while contrasting against negatives from different videos ($V_{2T1}$/$V_{2T2}$) and \textit{masking} potential false-negative pairs from the \textit{same} video ($V_{1T2}$), enforcing fine-grained discrimination (Eq.~\ref{eq:vtc-inst-loss}).}
 
    \label{fig:overall_framework}
\end{figure*}

\subsection{Self‑Supervised Masked Video Modeling}
\label{sec:video_mask}

Our method adopts a teacher-student framework to build our encoder, learning from semantic representations \cite{li2023unmasked}. While standard masked autoencoding with pixel-level reconstruction is data-efficient \cite{tong2022videomae, feichtenhofer2022masked}, this low-level objective can conflict with the high-level alignment needed for language tasks \cite{li2023unmasked, shu2022masked}. We therefore use a high-level feature regression on \textit{unmasked} tokens. This approach is significantly more training-efficient, as it removes the need for a heavy reconstruction decoder and saves considerable GPU memory by processing only the visible tokens \cite{li2023unmasked}. This semantic guidance also leads to faster convergence and produces representations that are better suited for subsequent cross-modal alignment \cite{li2023unmasked}.

Consider a video $\mathcal{V}=\{I_{1},\dots,I_{T}\}$ with $T$ RGB frames.
Each frame is divided into $N$ fixed‑size patches, producing a token sequence of length $L=T\!N$.
An \emph{attention‑guided} binary mask $\mathbf{M}\in\{0,1\}^{L}$ is constructed as follows.
A frozen vision transformer $g$ first processes all tokens to obtain self‑attention maps $\mathbf{A}\in\mathbb{R}^{L\times L}$.
Per-token importance scores are computed by averaging the attention \textbf{given} by each token,
\(\mathbf{s}= \tfrac{1}{L}\mathbf{A}\mathbf{1}.\)
Given a masking ratio $\rho$, the $L_{\mathrm{m}}=\lceil\rho L\rceil$ tokens with \emph{lowest} scores are masked ($\mathbf{M}=1$) while the remaining tokens are kept ($\mathbf{M}=0$).
Let $\Omega=\{l\mid\mathbf{M}_{l}=0\}$ be the visible index set.

A student video transformer $f_{\theta}$ receives only the visible tokens and outputs hidden vectors $\mathbf{h}^{S}_{l}$ for $l\in\Omega$.
The teacher features $\mathbf{h}^{T}_{l}=g(I_{1:T})_{l}$, computed on the \emph{full} token set, serve as regression targets.

The reconstruction loss is
\begin{equation}
\mathcal{L}_{\mathrm{rec}}\;=\;
\frac{1}{|\Omega|}\sum_{l\in\Omega}
\Bigl\lVert 
\frac{\mathbf{h}^{S}_{l}}{\lVert\mathbf{h}^{S}_{l}\rVert_{2}}
-
\frac{\mathbf{h}^{T}_{l}}{\lVert\mathbf{h}^{T}_{l}\rVert_{2}}
\Bigr\rVert_{2}^{2}
\label{eq:rec}
\end{equation}
This attention-guided masking compels the student to reconstruct the teacher's full-context representations (\(\mathbf{h}^T_l\)) for the most informative tokens (\(l \in \Omega\)), using only those same visible tokens as input. This challenging regression task strengthens its spatial-temporal representation.

\subsection{Instance-aware Global-Local Spatial-Temporal Alignment Learning }
\label{sec:multimodal}
% ----------------------------------------------------

Let \(\{(\mathcal{V}_{i},\mathcal{T}_{i})\}_{i=1}^{B}\) be paired video-text samples.
The visual encoder \(f_{\theta}\) (initialized from \S\ref{sec:video_mask}) yields token sequence \(\mathbf{V}_{i}\in\mathbb{R}^{L_{v}\times d}\) and pooled vector
\(
\mathbf{v}_{i}= \tfrac{1}{L_{v}}\sum_{l}\mathbf{V}_{i,l}.
\)
A text encoder~\cite{devlin2019bert} \(e_{\phi}\) outputs token embeddings \(\mathbf{T}_{i}\in\mathbb{R}^{L_{t}\times d}\) and pooled embedding
\(
\mathbf{t}_{i}= \mathbf{T}_{i,0},
\)
where the first token is the \texttt{[CLS]} representation.
Linear projections \(W_{v},W_{t}\in\mathbb{R}^{d\times d'}\) map pooled vectors to a shared space:
\(
\tilde{\mathbf{v}}_{i}=W_{v}\mathbf{v}_{i},\;
\tilde{\mathbf{t}}_{i}=W_{t}\mathbf{t}_{i}.
\)

\subsubsection{Global Alignment Losses}
With a learnable parameter temperature \(\tau\), the bidirectional Video-Text Contrastive (VTC) loss is
\begin{equation} % Use 'equation' for one equation number
\begin{split} % Use 'split' to break the line
\mathcal{L}_{\mathrm{VTC}} &=
-\frac{1}{B}\sum_{i=1}^{B}
\log\frac{\exp(\tilde{\mathbf{v}}_{i}^{\!\top}\tilde{\mathbf{t}}_{i}/\tau)}
{\sum_{j=1}^{B}\exp(\tilde{\mathbf{v}}_{i}^{\!\top}\tilde{\mathbf{t}}_{j}/\tau)} \\ % <-- Line break
&\quad \;-\;\frac{1}{B}\sum_{i=1}^{B} % <-- Align new line, indent with \quad
\log\frac{\exp(\tilde{\mathbf{t}}_{i}^{\!\top}\tilde{\mathbf{v}}_{i}/\tau)}
{\sum_{j=1}^{B}\exp(\tilde{\mathbf{t}}_{i}^{\!\top}\tilde{\mathbf{v}}_{j}/\tau)}
\end{split}
\end{equation}

A fusion transformer \(m_{\psi}\) (implemented as the BERT encoder) jointly processes the visual tokens \(\mathbf{V}_{i}\) and textual tokens \(\mathbf{T}_{i}\). A matching head \(h\) outputs logits from the fused \texttt{[CLS]} vector:
\(
s_{i}=h\!\bigl(m_{\psi}(\mathbf{V}_{i},\mathbf{T}_{i})\bigr)\in\mathbb{R}^{2}.
\)
Let \(y_{i}\in\{0,1\}\) indicate whether the pair is positive (1) or a hard negative (0).
With the softmax probability \(p_{i}=\operatorname{softmax}(s_{i})_{1}\), the binary cross‑entropy Video–Text Matching (VTM) loss is
\begin{equation}
\mathcal{L}_{\mathrm{VTM}}
=
-\frac{1}{B}\sum_{i=1}^{B}
\bigl[y_{i}\log p_{i}+(1-y_{i})\log(1-p_{i})\bigr]
\label{eq:vtm-loss}
\end{equation}

For each caption a subset \(M_{i}\subset\{1,\dots,L_{t}\}\) of token indices is replaced by \texttt{[MASK]}.
The Masked Language Modeling (MLM) loss, computed by the same fusion transformer $m_{\psi}$, is
\begin{equation}
\resizebox{0.9\linewidth}{!}{$
\mathcal{L}_{\mathrm{MLM}}
=
-\frac{1}{B}\sum_{i=1}^{B}\frac{1}{|M_{i}|}
\sum_{j\in M_{i}}\log P(w_{i,j}\mid\mathbf{V}_{i},\mathbf{T}_{i,\text{visible}})
$}
\label{eq:mlm-loss}
\end{equation}
where $P$ is the probability assigned by $m_{\psi}$ to the original word $w_{i,j}$.

\subsubsection{Instance‑Aware Alignment Losses}
\label{sec:instance}

Each video \(i\) is accompanied by \(K_{i}\) object instances described by bounding boxes \(b_{i,k}\) and captions \(\mathcal{T}_{i,k}\).
For every box, a crop \(\mathcal{C}_{i,k}\) is passed through the video encoder \(f_{\theta}\) to obtain:
(1) raw patch tokens \(\mathbf{C}_{i,k}\in\mathbb{R}^{L_{c}\times d}\) and 
(2) a raw pooled crop embedding $\mathbf{c}_{i,k} = \tfrac{1}{L_{c}}\sum_{l}\mathbf{C}_{i,k,l}$.

Cross‑attending the crop tokens to the full‑scene features \(\mathbf{V}_{i}\) injects global context:
\begin{align}
\mathbf{Z}_{i,k} &= \text{XAttn}(\mathbf{C}_{i,k},\mathbf{V}_{i}) \\
\mathbf{z}_{i,k} &= \frac{1}{L_{c}}\sum_{l=1}^{L_{c}}\mathbf{Z}_{i,k,l} \\
\tilde{\mathbf{z}}_{i,k} &= W_{v}\mathbf{z}_{i,k}
\end{align}
The text encoder returns a sentence embedding
\(
\mathbf{s}_{i,k}=e_{\phi}(\mathcal{T}_{i,k})_{\texttt{[CLS]}},
\;
\tilde{\mathbf{s}}_{i,k}=W_{t}\mathbf{s}_{i,k}.
\)

Since instance-level captions for objects within the same video/image often overlap (cf., Fig.~\ref{fig:ia_pt_dataset}) and can introduce false negatives in contrastive learning, we contrast each crop with all captions while masking non-matching captions from the same video/image, thereby promoting instance-level semantics.
With \(N=\sum_{i}K_{i}\) and an independent learnable temperature
\(\tau_{\mathrm{inst}}\), the instance VTC loss is:
\begin{equation}
\small
\begin{split}
\mathcal{L}_{\mathrm{VTC}}^{\mathrm{inst}} &=
-\frac{1}{N}\sum_{n=1}^{N}
\log
\frac{\exp\bigl(\tilde{\mathbf{z}}_{n}^{\!\top}\tilde{\mathbf{s}}_{n}/\tau_{\mathrm{inst}}\bigr)}
{\sum_{m=1}^{N}\alpha_{n,m}\,
\exp\bigl(\tilde{\mathbf{z}}_{n}^{\!\top}\tilde{\mathbf{s}}_{m}/\tau_{\mathrm{inst}}\bigr)} \\
&\quad\;-\;\frac{1}{N}\sum_{n=1}^{N}
\log
\frac{\exp\bigl(\tilde{\mathbf{s}}_{n}^{\!\top}\tilde{\mathbf{z}}_{n}/\tau_{\mathrm{inst}}\bigr)}
{\sum_{m=1}^{N}\alpha_{n,m}\,
\exp\bigl(\tilde{\mathbf{s}}_{n}^{\!\top}\tilde{\mathbf{z}}_{m}/\tau_{\mathrm{inst}}\bigr)}
\end{split}
\label{eq:vtc-inst-loss}
\end{equation}
where \(\alpha_{n,m}=0\) for \(m\neq n\) if \(m\) originates from the same video/image as \(n\), and \(\alpha_{n,m}=1\) otherwise.

The \emph{shared} fusion transformer $m_{\psi}$ and matching head $h$ are used for instance VTM. The model jointly encodes the raw pooled crop embedding $\mathbf{c}_{i,k}$ and the caption tokens of \(\mathcal{T}_{i,k}\), yielding logits \(s_{i,k}^{\mathrm{inst}}\in\mathbb{R}^{2}\).
The instance VTM objective trains the classifier to accept matched pairs and reject hard negatives:

\begin{equation}
\label{eq:vtm-inst-loss}
\begin{gathered}[t] % Stacks lines and top-aligns the number
\resizebox{0.9\columnwidth}{!}{$\displaystyle \mathcal{L}_{\mathrm{VTM}}^{\mathrm{inst}} = -\frac{1}{N}\sum_{i,k} \bigl[y_{i,k}\log p_{i,k}+(1-y_{i,k})\log(1-p_{i,k})\bigr]$} \\
p_{i,k} = \operatorname{softmax}\bigl(s_{i,k}^{\mathrm{inst}}\bigr)_{1}
\end{gathered}
\end{equation}

Similar to the global MLM loss, we randomly mask a subset \(M_{i,k}\) of caption tokens and ask the \emph{shared} fusion transformer $m_{\psi}$ to recover them, but this time given the cross-attended visual context $\mathbf{Z}_{i,k}$:
\begin{equation}
\label{eq:mlm-inst-loss}
\resizebox{0.88\columnwidth}{!}{%  <-- Your 70% scaling
  $\displaystyle                  %  <-- Keep math style
  \begin{aligned}[t] %  <-- Top-align the equation number
  \mathcal{L}_{\mathrm{MLM}}^{\mathrm{inst}}
  &= -\frac{1}{N}\sum_{i,k} \frac{1}{|M_{i,k}|} \sum_{j\in M_{i,k}} \log P\bigl(w_{i,k,j}\mid \mathbf{Z}_{i,k}, 
  \mathcal{T}_{i,k,\text{visible}}\bigr) % <-- Align this line
  \end{aligned}
  $
}
\end{equation}

% --- This final section was already correct ---
Combining the masked-video alignment with the three
pair-level objectives (\(\mathcal{L}_{\mathrm{VTC}},\mathcal{L}_{\mathrm{VTM}},\mathcal{L}_{\mathrm{MLM}}\))
and their instance-aware counterparts yield our complete training loss.
We introduce separate weighting coefficients so that each component can
be tuned independently, leading to the following decomposition.

\begin{align}
\mathcal{L}_{\mathrm{global}}
  &= \lambda_{\mathrm{VTC}}\,\mathcal{L}_{\mathrm{VTC}}
  + \lambda_{\mathrm{VTM}}\,\mathcal{L}_{\mathrm{VTM}} \nonumber \\ % \nonumber removes number for this line
  &\qquad + \lambda_{\mathrm{MLM}}\,\mathcal{L}_{\mathrm{MLM}} \label{eq:global-loss} \\
\mathcal{L}_{\mathrm{inst}}
  &= \lambda_{\mathrm{VTC}}^{\mathrm{inst}}\,\mathcal{L}_{\mathrm{VTC}}^{\mathrm{inst}}
  + \lambda_{\mathrm{VTM}}^{\mathrm{inst}}\,\mathcal{L}_{\mathrm{VTM}}^{\mathrm{inst}} \nonumber \\
  &\qquad + \lambda_{\mathrm{MLM}}^{\mathrm{inst}}\,\mathcal{L}_{\mathrm{MLM}}^{\mathrm{inst}} \label{eq:inst-loss}
\end{align}

The complete loss integrates masked‑video reconstruction, global
video-text alignment, and the three instance‑level objectives:
\begin{equation}
\mathcal{L}
=
\mathcal{L}_{\mathrm{rec}}
+\mathcal{L}_{\mathrm{global}}
+\mathcal{L}_{\mathrm{inst}}
\label{eq:total-loss}
\end{equation}
\section{Experimental Setup}
\label{sec:exp_setup}
% ----------------------------------------------------

We use a Vision Transformer Large (ViT-L)~\cite{dosovitskiy2020image} trained from scratch but guided by a frozen original CLIP-ViT teacher~\cite{radford2021learning}. 
While several OpenCLIP~\cite{cherti2023reproducible} models have shown strong performance on standard benchmarks, in our experiments we found the original CLIP-ViT-L teacher to provide a stronger signal, as the OpenCLIP variants performed worse even at higher native resolutions (e.g., $378 \times 378$)~\cite{fang2023data}. This observation aligns with recent findings in the development of vision
encoders for multimodal learning~\cite{Li_Open_2025_ICCV}. Following the strategy in~\cite{li2023unmasked}, the class token is removed and all patch tokens attend jointly in space and time. This preserves the teacher’s spatial semantics while enabling explicit spatial-temporal reasoning in the student.

\subsection{Self‑Supervised Masked Video Modeling}
\label{sec:stage1}

The model was pretrained for $800$ epochs on $8$-frame \(224\times224\) video clips, using only videos from three corpora: K710 (\(0.6\)M videos), segmented HDVILA ($0.45$M videos), and WebVid ($0.45$M videos). We merge Kinetics-400, -600, and -700~\cite{kay2017kinetics} into \emph{Kinetics-710}; due to YouTube removals, approximately $15\%$ of the videos are missing. We use AdamW~\cite{loshchilov2017decoupled} optimizer with a learning rate of \(1.5\times10^{-4}\) and a batch size of $64$, alongside an $80\%$ attention-guided masking ratio.

After pretraining, we select a set of checkpoints with the lowest alignment loss between teacher and student. For each checkpoint, we append a linear classifier and fine-tune the entire network on Kinetics-400 for action classification. Among these candidates, we choose the model achieving the highest Top-1 accuracy on Kinetics-400 ($87.84\%$ top-1, $97.77\%$ top-5), and use corresponding pre-trained weights for continued instance-aware alignment training. This first pre-training stage was run on $320$ NVIDIA H100 GPUs.

\subsection{Instance-aware Alignment Learning}
\label{sec:stage2}

We use a large collection of image-text pairs including CC3M~\cite{sharma2018conceptual}, CC12M~\cite{changpinyo2021conceptual}, SBU Captions~\cite{ordonez2011im2text}, Visual Genome~\cite{krishna2017visual}, COCO~\cite{lin2014microsoft}, and ShareGPT4V~\cite{chen2024sharegpt4v}, alongside $5$ million sampled WebVid~\cite{bain2021frozen} videos for global alignment. Our InstVL training set of $2$ million images and $50,000$ videos is used for both global and instance-aware alignment.

Initializing the vision encoder with weights from masked video modeling, we train on a mixture of image-text and video-text pairs for $15$ epochs. We conducted experiments sampling $4$, $8$, $16$, $24$, and $32$ frames, finding that $16$ frames yielded the best performance, while $32$ frames showed a slight degradation. Therefore, we sample $16$ frames per video at $224\times224$, and still images are treated as single-frame videos. Because InstVL captions often exceed the tokenizer’s input length, at each epoch we randomly sample one sentence per caption, cycling through all sentences across epochs so the model eventually sees every part of each description. Ablations in Table~\ref{tab:ablation} analyze the impact of sampling strategy.

Zero‑shot retrieval is assessed on MSVD~\cite{chen2011collecting}, ActivityNet~\cite{caba2015activitynet}, MSR-VTT~\cite{xu2016msr}, LSMDC~\cite{rohrbach2017movie}, DiDeMo~\cite{anne2017localizing}, and InstVL test sets without additional fine‑tuning. This second alignment stage was trained on $200$ NVIDIA B200 GPUs with $180$GB of memory per GPU. We use the AdamW optimizer~\cite{loshchilov2017decoupled} with a cosine learning scheduler.

\section{Results}
\label{sec:results}

\begin{table*}
  \centering
  \caption{Comparison of SOTA models and our InstAP on the InstVL test set. We report T2V/V2T R@1 on the \emph{instance} and \emph{global} splits across InstVL(\texttt{img}), InstVL(\texttt{img-zero}), and InstVL(\texttt{video}). \texttt{UMT-L (InstVL; g/g+i)} baselines use the same full training corpus as InstAP, trained with only InstVL's global captions (g) or with all InstVL captions treated as global (g+i).}

  \label{tab:instvl_combined}
  \resizebox{0.88\textwidth}{!}{%
    \begin{tabular}{@{} ll
                      cc cc   % InstVL(img): 1K & 10K
                      cc cc   % InstVL(img-zero): 1K & 10K
                      cc      % InstVL(video): only 1K
                    @{}}
      \toprule
     
      \multirow{3}{*}{Method} & \multirow{3}{*}{Split}
      & \multicolumn{4}{c}{InstVL(img)}
      & \multicolumn{4}{c}{InstVL(img-zero)}
      & \multicolumn{2}{c}{InstVL(video)} \\
      \cmidrule(lr){3-6}\cmidrule(lr){7-10}\cmidrule(lr){11-12}
     
      & 
      & \multicolumn{2}{c}{1K} & \multicolumn{2}{c}{10K}
      & \multicolumn{2}{c}{1K} & \multicolumn{2}{c}{10K}
      & \multicolumn{2}{c}{1K} \\
      \cmidrule(lr){3-4}\cmidrule(lr){5-6}
      \cmidrule(lr){7-8}\cmidrule(lr){9-10}
      \cmidrule(lr){11-12}
    
      &
      & T2V R@1 & V2T R@1 & T2V R@1 & V2T R@1
      & T2V R@1 & V2T R@1 & T2V R@1 & V2T R@1
      & T2V R@1 & V2T R@1 \\
      \midrule

      \multirow{2}{*}{VideoPrism~\cite{zhao2024videoprism}} & Instance
      & 28.21 & 34.52 & 22.75 & 29.51 & 21.32 & 27.39 & 13.85 & 20.04 & 40.86 & 39.29 \\
      & Global
      & 97.40 & 97.60 & 88.19 & 89.62 & 85.70 & 85.80 & 73.05 & 75.11 & 82.71 & 83.62 \\
      \multirow{2}{*}{CLIP4Clip~\cite{luo2022clip4clip}} & Instance
      & 25.10 & 33.21  & 18.68 & 28.19  & 17.82 & 25.10 & 9.11 & 16.30 & 17.71 & 24.69\\
      & Global
      & 93.40 & 96.00  & 79.22 & 84.25  & 78.20 & 81.70 & 56.95 & 63.96 & 67.50 & 70.50\\      
      \multirow{2}{*}{Coca~\cite{yu2022coca}} & Instance
      & 11.83 & 21.79 & 7.36 & 13.33 & 7.08 & 13.19 & 4.12 & 7.26 & 14.72 & 11.82 \\
      & Global
      & 86.20 & 91.50 & 70.80 & 76.16 & 67.40 & 70.50 & 46.05 & 50.64 & 46.92 & 43.78 \\      
      \multirow{2}{*}{ViCLIP~\cite{wang2023internvid}} & Instance
      & 28.38 & 28.91 & 19.46 & 20.02 & 18.25 & 20.93 & 9.57 & 11.21 & 21.78 & 21.50 \\
      & Global
      & 95.10 & 93.50 & 81.47 & 79.33 & 77.80 & 77.60 & 58.51 & 58.21 & 62.89 & 62.69 \\      
      \multirow{2}{*}{OpenCLIP~\cite{cherti2023reproducible}} & Instance
      & 37.88 & 44.06 & 29.21 & 37.76 & 26.73 & 36.19 & 17.28 & 25.57  & 36.63 & 33.36 \\
      & Global
      & 94.40 & 98.10 & 84.98 & 92.06 & 83.40 & 86.90 & 70.75 & 78.13 & 82.00 & 77.15 \\
      \multirow{2}{*}{CLIP-ViP~\cite{xue2023clip}} & Instance
      & 24.04 & 32.06 & 14.38 & 21.85 & 13.81 & 22.96 & 6.60 & 12.11  & 16.78 & 28.32 \\
      & Global
      & 78.40 & 89.20 & 54.94 & 72.00 & 55.60 & 73.20 & 32.48 & 51.30 & 35.59 & 61.07 \\
      \multirow{2}{*}{MCQ~\cite{ge2022bridging}} & Instance
      & 19.33 & 22.11 & 9.63 & 11.13 & 17.08 & 19.61 & 7.04 & 8.55  & 24.41 & 23.72 \\
      & Global
      & 58.20 & 60.10 & 31.45 & 34.12 & 58.90 & 62.70 & 34.13 & 38.26 & 61.48 & 60.67 \\
      \multirow{2}{*}{SigLIP~\cite{zhai2023sigmoid}} & Instance
      & 38.17 & 45.17 & 29.76 & 37.83 & 28.25 & 35.56 & 16.98 & 25.19  & 36.43 & 36.14 \\
      & Global
      & 95.70 & 98.20 & 87.18 & 91.97 & 83.90 & 86.50 & 68.64 & 75.66 & 74.72 & 76.14 \\
      \multirow{2}{*}{UMT-L~\cite{li2023unmasked}} & Instance
      & 38.44 & 35.65 & 21.34 & 23.08 & 29.34 & 30.17 & 11.09 & 16.38 & 26.38 & 22.43 \\
      & Global
      & 94.70 & 95.30 & 83.95 & 85.41 & 83.90 & 83.70 & 72.60 & 72.59 & 88.30 & 85.50 \\
      \midrule
      \multirow{2}{*}{UMT-L (InstVL; g)~\cite{li2023unmasked}} & Instance
      & 34.44 & 41.24 & 22.87 & 30.37
      & 25.97 & 31.97 & 13.33 & 19.21
      & 41.51 & 40.34 \\
      & Global
       & 96.20 & 97.10 & 85.70 & 87.03
      & 85.30 & 86.40 & 72.50 & 74.18 
      & 84.80  & 82.40 \\
      \multirow{2}{*}{UMT-L (InstVL; g+i)~\cite{li2023unmasked}} & Instance
      & 45.74 & 44.27 & 34.83 & 35.15 
      & 34.68 & 34.99  & 21.13 & 22.82
      & 40.38 & 39.33 \\
      & Global
       & 93.20  & 94.30 & 80.30 & 81.62
      & 82.40    & 84.30   & 68.16 & 69.76 
      & 79.90 & 77.20 \\
      \multirow{2}{*}{\textbf{InstAP (Ours)}} & Instance
      & \textbf{50.25} & \textbf{49.26} & \textbf{44.05} & \textbf{45.76}
      & \textbf{41.94} & \textbf{42.53} & \textbf{28.25} & \textbf{31.87}
      & \textbf{60.63} & \textbf{58.49} \\
      & Global
      & \textbf{99.20} & \textbf{99.10} & \textbf{95.77} & \textbf{94.71}
      & \textbf{88.70} & \textbf{88.30} & \textbf{83.33} & \textbf{82.21}
      & \textbf{94.50} & \textbf{95.50} \\
      \bottomrule
    \end{tabular}%
  }
\end{table*}

Table~\ref{tab:instvl_combined} compares InstAP against state-of-the-art models on InstVL benchmarks. For fair comparison in instance-level tasks, baselines are evaluated using cropped regions/trajectories, which consistently yielded stronger results than full-frame inputs. InstAP achieves superior performance across all image and video splits for both instance and global retrieval. Notably, on \texttt{InstVL-1K (video)} instance retrieval, InstAP reaches $60.63$ T2V R@1, significantly exceeding prior work. Strong performance on the unseen \texttt{img-zero} splits further suggests generalization beyond training data memorization.

To isolate the benefits of our framework from the InstVL dataset itself, we compare InstAP against two UMT-L baselines trained on the same corpus: (1) \texttt{UMT-L (g)}, using only global captions; and (2) \texttt{UMT-L (g+i)}, using both global and instance captions as standard global-level descriptions. InstAP significantly outperforms \texttt{UMT-L (g+i)} (e.g., $44.05$ vs. $34.83$ T2V R@1 on \texttt{InstVL-10K (img)}), despite identical training data. This gap confirms that InstAP's gains are driven by our novel instance-aware alignment framework rather than mere exposure to dense annotations.

\begin{table*}[t]
    \centering
    \small              % slightly smaller font
    \setlength{\tabcolsep}{3pt} % reduce column padding
    \caption{Zero-shot text-to-video retrieval (R@1 / R@5 / R@10) on standard benchmarks. \texttt{UMT-L (InstVL; g)} and \texttt{UMT-L (InstVL; g+i)} are baselines trained on the full corpus as InstAP.}
    \label{tab:instvl_zeroshot}

    \resizebox{0.61\textwidth}{!}{%
    \begin{tabular}{lccccc}
        \toprule
        \textbf{Method} & \textbf{MSR-VTT} & \textbf{DiDeMo} & \textbf{MSVD} & \textbf{LSMDC} & \textbf{ActivityNet} \\
        \midrule
        CLIP4Clip~\cite{luo2022clip4clip} & 32.0 / 57.0 / 66.9 & -- & 38.5 / 66.9 / 76.8 & 15.1 / 28.5 / 36.4 & -- \\
        Frozen in Time~\cite{bain2021frozen}         & 18.7 / 39.5 / 51.6 & 21.1 / 46.0 / 56.2 & 38.7 / 70.1 / 80.1 & 9.3 / 22.0 / 30.1 & -- \\
        VIOLET~\cite{fu2021violet}          & 25.9 / 49.5 / 59.7 & 23.5 / 49.8 / 59.8 & -- & -- & -- \\
        ALPRO~\cite{li2022align}           & 24.1 / 44.7 / 55.4 & 23.8 / 47.3 / 57.9 & -- & -- & -- \\
        RAP~\cite{wu2022rap}             & 28.9 / 47.5 / 56.8 & 29.5 / 55.7 / 65.6 & 35.9 / 64.3 / 73.7 & 12.8 / 26.6 / 33.4 & -- \\
        Clover~\cite{huang2023clover}          & 26.4 / 49.5 / 60.0 & 29.5 / 55.2 / 66.3 & -- & 14.7 / 29.2 / 38.2 & -- \\
        TW-BERT~\cite{yang2023learning}         & 26.4 / 50.1 / 59.6 & 28.4 / 52.9 / 64.5 & -- & 14.2 / 30.4 / 36.0 & -- \\
        Singularity~\cite{lei2023revealing}     & 28.4 / 50.2 / 59.5 & 36.9 / 52.9 / 64.5 & -- & -- & -- \\
        LaT~\cite{bai2022lat}             & 23.4 / 44.1 / 53.3 & 22.6 / 45.9 / 58.9 & 36.9 / 68.6 / 81.0 & -- & -- \\
        OA-Trans~\cite{wang2022object}        & 23.4 / 47.5 / 55.6 & 23.5 / 50.4 / 59.8 & -- & -- & -- \\
        MCQ~\cite{ge2022bridging}             & 26.0 / 46.4 / 56.4 & 25.6 / 50.6 / 61.1 & 43.6 / 74.9 / 84.9 & 12.2 / 25.9 / 32.2 & -- \\
        MILES~\cite{ge2022miles}           & 26.1 / 47.2 / 56.9 & 27.2 / 50.3 / 63.6 & 44.4 / 76.2 / \textbf{87.0} & 11.1 / 24.7 / 30.6 & -- \\
        CLIP-ViP~\cite{xue2023clip}        & 31.7 / 51.2 / 63.2 & 24.6 / 50.7 / 59.7 & -- & 12.5 / 26.1 / 33.3 & -- \\
        EA-VTR~\cite{ma2024ea}          & 28.0 / 53.1 / 62.3 & 32.7 / 58.9 / 68.9 & 46.6 / \textbf{78.9} / 86.5 & 15.7 / 29.6 / 36.0 & -- \\
        UMT-L~\cite{li2023unmasked}             & 39.7  /  61.8  /  70.9
                        & 47.0  / 71.8  /  78.8 
                        & 47.0 /  75.4  /  83.6  
                        & \textbf{26.0} /  \textbf{43.1}  / \textbf{51.6}
                        & 44.3 /  72.2 /  84.4 \\
        \midrule
        UMT-L (InstVL; g)~\cite{li2023unmasked}
                        & 35.4 / 59.4 / 70.2
                        & 44.1 / 72.3 / 79.1
                        & 43.7  / 73.4 / 82.4
                        & 19.9  / 38.4 / 46.5
                        & 39.8 / 66.5 / 76.5 \\
        UMT-L (InstVL; g+i)~\cite{li2023unmasked}
                        & 34.0  / 58.5 / 68.5
                        & 42.7 / 69.0 / 77.0
                        & 41.3 /  71.8 /   81.4
                        & 17.5 / 36.6 / 46.5
                        & 37.1 / 64.5 / 74.7 \\
        \textbf{InstAP (Ours)}  & \textbf{41.1} / \textbf{65.2} /   \textbf{73.6}
                        &  \textbf{54.0}  / \textbf{78.2}  /  \textbf{84.5}
                        & \textbf{49.2}  / 77.0 /  85.1 
                        & 23.5  / 42.7 /   50.3 
                        & \textbf{50.7} / \textbf{77.2} /  \textbf{86.6} \\
        \bottomrule
    \end{tabular}
    }
\end{table*}

Table~\ref{tab:instvl_zeroshot} evaluates InstAP's generalization across five zero-shot text-to-video retrieval benchmarks. InstAP reaches $41.1$ R@1 on MSR-VTT and $54.0$ on DiDeMo, setting new state-of-the-art performance levels. Crucially, we observe that naively fine-tuning the UMT-L baseline on InstVL (\texttt{g} or \texttt{g+i} variants) degrades performance compared to the original UMT-L, likely due to task interference or domain shift. In contrast, InstAP not only mitigates this degradation but surpasses the original UMT-L on both MSR-VTT and DiDeMo while remaining competitive elsewhere. This demonstrates that our instance-aware paradigm fosters more robust, dual-granularity representations that benefit both fine-grained grounding and global understanding.

To further validate the instance-awareness of our representations beyond retrieval, we evaluate visual grounding on the \texttt{InstVL-1K} splits. We attach a $3$-layer MLP box-regression head to the fused vision-text features of the pre-trained encoder and fine-tune using L1 and GIoU losses. As shown in Table~\ref{tab:instvl_grounding}, InstAP significantly outperforms the UMT-L~\cite{li2023unmasked} baseline across all datasets and IoU thresholds. Notably, on the challenging \texttt{video} split, InstAP improves IoU@$90$ from $14.44$ to $25.13$, confirming that our pre-training objective effectively encodes precise spatial-temporal coordinates within the visual features.

\begin{table}[h]
  \centering
  \caption{Grounding metrics (IoU@\{50, 70, 90\}) on \texttt{InstVL-1K}.}
  \label{tab:instvl_grounding}
  \resizebox{\columnwidth}{!}{%
    \begin{tabular}{@{} l ccc ccc ccc @{}}
      \toprule
      \multirow{2}{*}{Method} & \multicolumn{3}{c}{InstVL(img)} & \multicolumn{3}{c}{InstVL(img-zero)} & \multicolumn{3}{c}{InstVL(video)} \\
      \cmidrule(lr){2-4}\cmidrule(lr){5-7}\cmidrule(l){8-10} % <--- Trimming removed from the right here
      & IoU@50 & IoU@70 & IoU@90 & IoU@50 & IoU@70 & IoU@90 & IoU@50 & IoU@70 & IoU@90 \\
      \midrule
      UMT-L & 74.53 & 63.47 & 41.64  & 67.12 & 54.20 & 34.05  & 54.25 & 40.70 & 14.44 \\
      \textbf{InstAP (Ours)} & \textbf{76.17} & \textbf{67.04} & \textbf{48.20} & \textbf{68.52} & \textbf{58.91} & \textbf{42.14} & \textbf{60.02} & \textbf{48.85} & \textbf{25.13} \\
      \bottomrule
    \end{tabular}%
  }
\end{table}

\begin{table}
\centering
\caption{Effect of adding the instance-aware loss $\mathcal{L}_{\mathrm{inst}}$ to the base objectives $\mathcal{L}_{\mathrm{rec}}+\mathcal{L}_{\mathrm{global}}$. We report mean recall (average of R@$1$, R@$5$, R@$10$ over T2V and V2T) on standard and InstVL benchmarks.}

\label{tab:compare_global_pub_modified} 
\resizebox{\columnwidth}{!}{%
\begin{tabular}{lccccccc}
\toprule
\multirow{2}{*}{Alignment} & \multirow{2}{*}{DiDeMo} & \multirow{2}{*}{MSR-VTT} & \multirow{2}{*}{LSMDC} & \multicolumn{2}{c}{InstVL-1K (img-zero)} & \multicolumn{2}{c}{InstVL-1K (video)} \\ \cmidrule(lr){5-6} \cmidrule(lr){7-8}
& & & & Instance & Global & Instance & Global \\
\midrule
$\mathcal{L}_{\mathrm{rec}} + \mathcal{L}_{\mathrm{global}}$
& 65.98 & 54.65 & 34.47 & 49.98 & 88.82 & 57.71 & 91.55 \\
$\mathcal{L}_{\mathrm{rec}} + \mathcal{L}_{\mathrm{global}} + \mathcal{L}_{\mathrm{inst}}$
& \textbf{70.01} & \textbf{56.72} & \textbf{35.75} & \textbf{63.94} & \textbf{89.78} & \textbf{75.32}  & \textbf{97.03}  \\
\bottomrule
\end{tabular}
}
\end{table}

To investigate the individual contribution of our proposed instance-aware alignment loss ($\mathcal{L}_{\mathrm{inst}}$), we conduct a detailed ablation study presented in Table~\ref{tab:compare_global_pub_modified}. We compare our full InstAP model, which utilizes all objectives ($\mathcal{L}_{\mathrm{rec}} + \mathcal{L}_{\mathrm{global}} + \mathcal{L}_{\mathrm{inst}}$), against a variant trained with only reconstruction and global alignment ($\mathcal{L}_{\mathrm{rec}} + \mathcal{L}_{\mathrm{global}}$). The results are conclusive: the addition of $\mathcal{L}_{\mathrm{inst}}$ is the critical component for fine-grained understanding. It provides a massive boost to instance-level retrieval, improving the mean recall on the \texttt{InstVL-1K (video)} instance split from $57.71$ to $75.32$ ($+17.61$) and on the \texttt{InstVL-1K (img-zero)} instance split from $49.98$ to $63.94$ ($+13.96$). This demonstrates that global alignment alone is insufficient for this challenging task. Furthermore, this focus on fine-grained details does not come at the cost of global understanding; it significantly enhances it. The full model with $\mathcal{L}_{\mathrm{inst}}$ also achieves the best performance on all global-only benchmarks, including \texttt{InstVL-1K (video)} global ($97.03$ vs. $91.55$) and standard datasets like DiDeMo ($70.01$ vs. $65.98$). This confirms that $\mathcal{L}_{\mathrm{inst}}$ is essential for instance-level capabilities and simultaneously improves the robustness of the global representations.

\begin{table}
  \centering
  \caption{Ablation of InstAP components on the InstVL instance-level test sets. We report mean recall, averaged over R@$1$, R@$5$, and R@$10$ for both V2T and T2V retrieval.}
  \label{tab:ablation}
  \resizebox{\linewidth}{!}{%
    \begin{tabular}{lccc}
      \toprule
      Method                         & InstVL-1K (img) & InstVL-1K (img-zero) & InstVL-1K (video)\\
      \midrule
      Baseline                       &    59.10      &   46.37      &    45.48         \\
      + Instance temperature         &    67.19        &  54.90       &   55.22              \\
      + Weighted instance loss           &    68.17      &  56.00       &   58.16            \\
      + Caption sub-sampling         &   71.65       &  58.42       &    58.97            \\
      + \textbf{Instance trajectory}         &  \textbf{75.03}         &  \textbf{63.94}       &    \textbf{75.32}            \\
      \bottomrule
    \end{tabular}%
  }
\end{table}

We ablate the components of InstAP in Table~\ref{tab:ablation}, showing cumulative gains over a baseline that already includes $\mathcal{L}_{\mathrm{inst}}$. First, a learnable instance temperature yields a substantial improvement (e.g., $+8.09$ on \texttt{InstVL-1K (img)}). Second, weighting the instance loss ($\lambda^{\mathrm{inst}} = 0.1$) provides a consistent gain by better balancing the sparse instance data within the large-scale training mixture. Third, caption sub-sampling serves as an effective regularizer for InstVL's long descriptions and brings further improvement. Finally, adding the $50$K video trajectory dataset gives the largest boost ($+16.35$ on \texttt{InstVL-1K (video)}), highlighting that explicit pre-training on temporal trajectories is critical for spatial-temporal understanding.

\begin{figure}
    \centering
    \includegraphics[width=\linewidth]{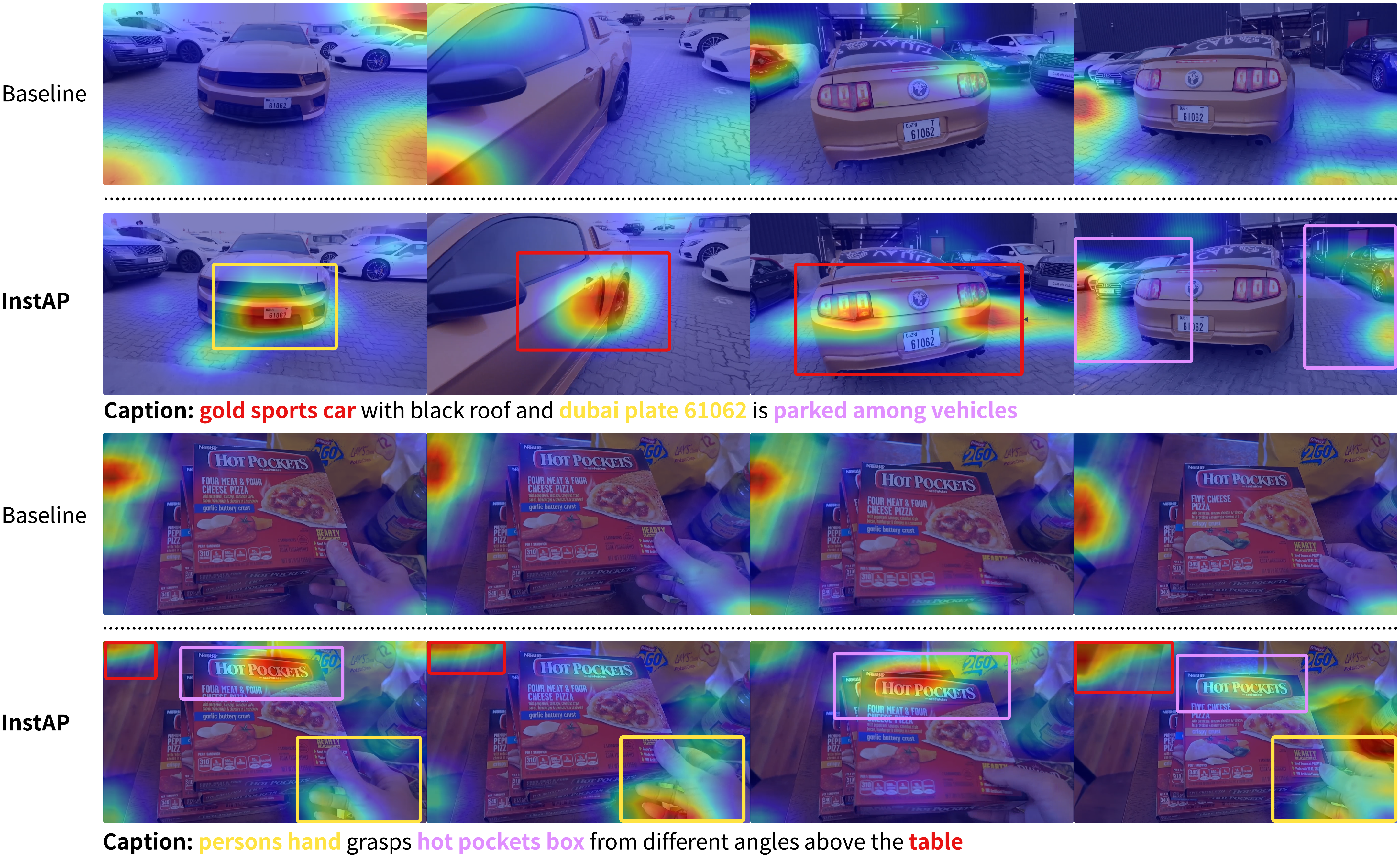}
    \caption{InstAP tends to attend more closely to caption-relevant regions (e.g., `dubai plate 61062') than the global-only baseline~\cite{li2023unmasked}, which often exhibits diffuse or misaligned attention.}
    \label{fig:viz_attn}
\end{figure}

\begin{figure}
    \centering
    \includegraphics[width=0.94\linewidth]{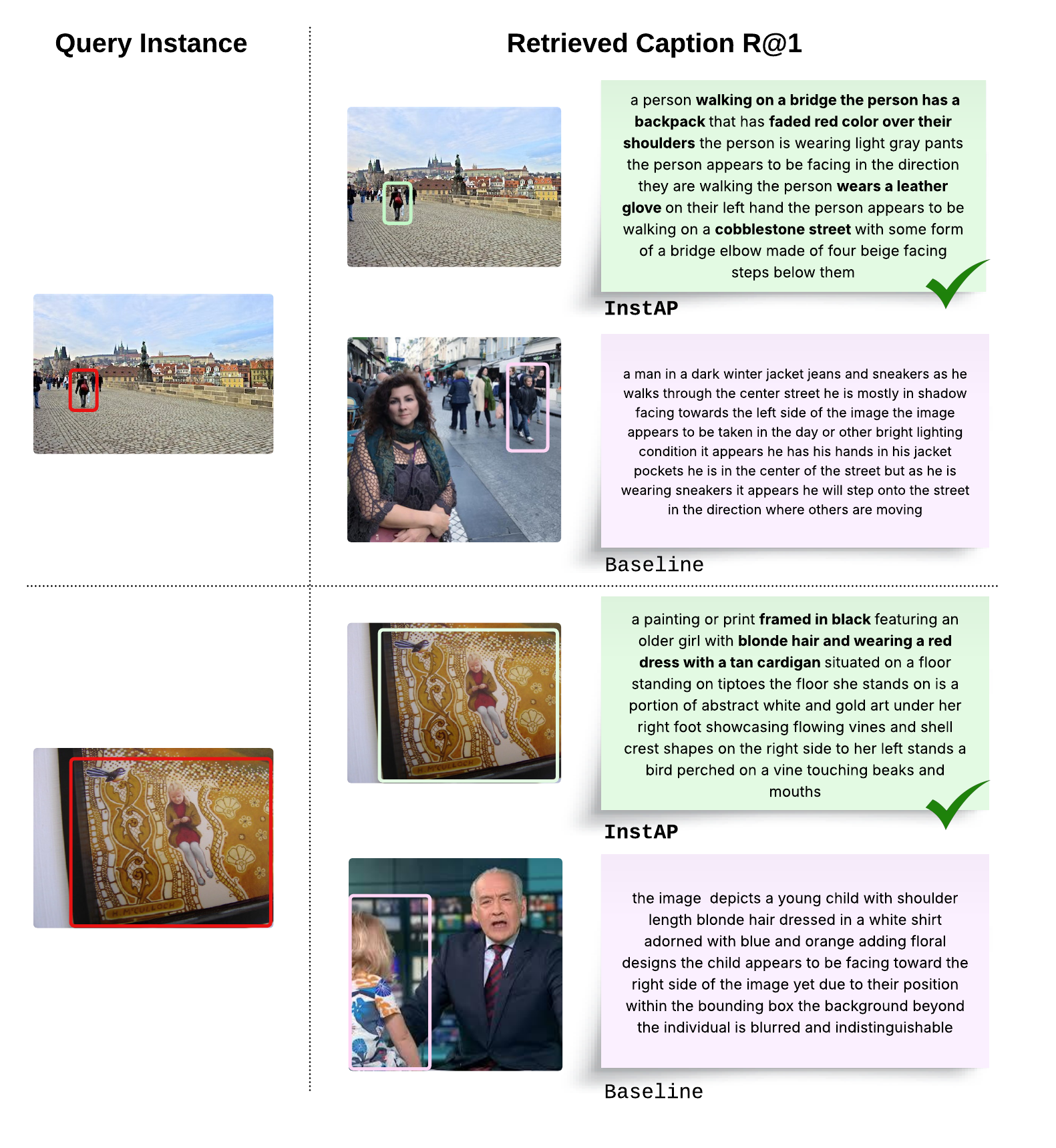}
    \caption{InstAP consistently retrieves correct fine-grained descriptions, whereas the global baseline~\cite{li2023unmasked} is confounded by semantic distractors and mismatches the query.}
    \label{fig:failure_cases}
\end{figure}

Figure~\ref{fig:viz_attn} visualizes InstAP's grounding capabilities using gradient-weighted activation mapping with rank-based Gaussian filtering~\cite{Li_2025_ICCV}. While baseline attention is typically diffuse, InstAP precisely localizes textual phrases to specific spatial-temporal regions. This superior grounding translates to more accurate instance retrieval, as illustrated in Fig.~\ref{fig:failure_cases}. 

Our analysis of $1{,}500$ instance-retrieval errors identifies the top three failure modes as multi-instance confusion under heavy occlusion or clutter at $44.6\%$, limited visual evidence in background-dominant or small-scale crops at $24.6\%$, and cross-sample semantic matches at $13.1\%$. Together, these account for $82.3\%$ of all errors, indicating that clutter and sparse visual signals remain key challenges.

\section{Conclusion}
\label{conclusion}
We introduce InstAP, an instance-aware pre-training framework for fine-grained video-language understanding. Built on the large-scale InstVL dataset with dual-granularity annotations, InstAP learns to ground text in specific spatial-temporal trajectories through an instance-aware alignment objective. Experiments show that its gains come from the training paradigm rather than from data alone, as it consistently outperforms strong baselines trained on the same dataset. Importantly, this instance-level pre-training also improves global representations, leading to strong generalization across standard benchmarks. Overall, InstAP advances VLP models toward more robust understanding of complex visual scenes at both holistic and instance levels.

\section*{Acknowledgment}
{
This work was supported by project JPNP20017, which was subsidized by the New Energy and Industrial Technology Development Organization (NEDO).
}
{
    \small
    \bibliographystyle{ieeenat_fullname}
    \bibliography{main}
}


\end{document}